\documentclass[10pt,twocolumn,letterpaper]{article}

\usepackage[pagenumbers]{cvpr} 
\definecolor{cvprblue}{rgb}{0.21,0.49,0.74}
\usepackage[pagebackref,breaklinks,colorlinks,allcolors=cvprblue]{hyperref}

\usepackage{xcolor} 

\usepackage{algorithm}
\usepackage{algorithmic}
\usepackage{makecell} 
\usepackage{tikz}
\usepackage{pifont}

\newcommand{\machine}{%
  \tikz[baseline=-0.6ex]{%
    \fill[green!70!black] (0,0) circle (0.95ex);
    \node[text=white,font=\sffamily\bfseries,scale=0.9] at (0,0){\ding{52}}; 
  }%
}
\newcommand{\manual}{%
  \tikz[baseline=-0.6ex]{%
    \draw[red!75!black, thick] (0,0) circle (0.95ex);
    \node[text=red!75!black,font=\sffamily\bfseries,scale=0.9] at (0,0){\ding{55}}; 
  }%
}

\newsavebox{\partialmanualbox}
\savebox{\partialmanualbox}{%
  \tikz[baseline=-0.6ex]{%
    \draw[yellow!75!black, thick] (0,0) circle (0.95ex);
    \node[text=yellow!75!black,font=\sffamily\bfseries,scale=0.9] at (0,0){\ding{51}}; 
  }%
}
\newcommand{\partialmanual}{\usebox{\partialmanualbox}}

\usepackage{multirow}
\usepackage{makecell} 

\def\paperID{4} 
\def\confName{CVPR}
\def\confYear{2026}

\title{USAI-Quant: A Quantitative Reasoning Benchmark for Vision-Language Models in Built Environments}

\author{Dongdong Wang\thanks{\raggedright Corresponding authors: \texttt{\{dongdongwang@ufl.edu, shenhaowang@ufl.edu}}\\
University of Florida\\
\and
Qingqi Song\\
University of Florida\\
\and
Yuzhou Chen\\
University of Florida\\
\and
Deepak Balakrishnan\\
University of Florida\\
\and
Ravi Shankar Srinivasan \\
University of Florida\\
\and
Shenhao Wang\footnotemark[1]\\
University of Florida
}

\begin{document}

\maketitle

\begin{figure*}[h]
    \centering
    \includegraphics[width=1\linewidth]{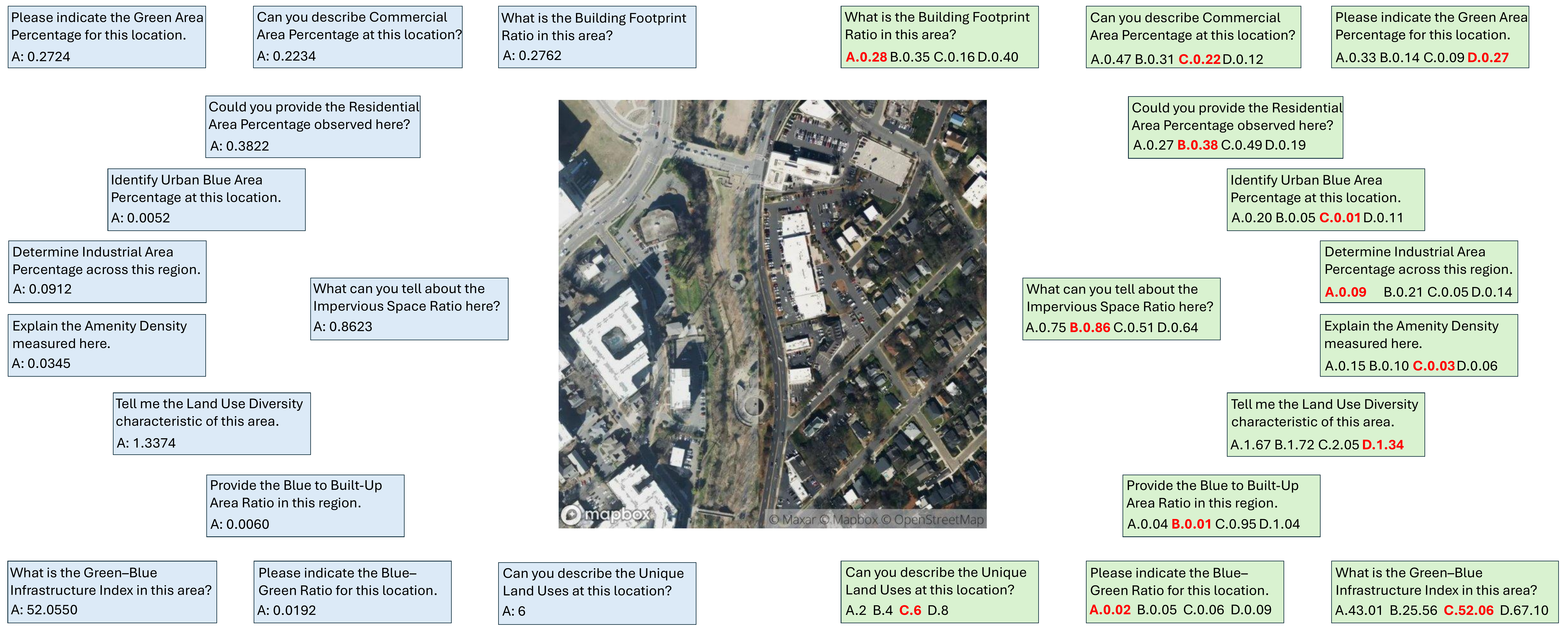}
    \caption{An example from USAI-Quant. USAI-Quant is designed to evaluate quantitative reasoning for built environment metrics through remote sensing imagery. It integrates 13 urban metrics and supports both open-ended and multiple-choice question formats, enabling comprehensive assessment of quantitative VQAs.}
    \label{fig:vqa_illustration}
\end{figure*}
\begin{abstract}

Large vision–language models (VLMs) have emerged as a powerful paradigm for urban and spatial AI. However, current state-of-the-art large VLMs still struggle with quantitative reasoning on remote sensing imagery. Existing benchmarks and algorithms are predominantly based on qualitative Visual Question Answering (VQA), providing limited insights into the quantitative reasoning capabilities of VLMs for built environment metrics. To address this gap, we develop \textbf{Quant}itative \textbf{U}rban and \textbf{S}patial \textbf{AI} benchmark (\textbf{USAI-Quant}), the first benchmark designed to quantitatively evaluate VLM's reasoning capabilities on built environment metrics via remote sensing imagery. USAI-Quant is curated from the 335 largest U.S. cities, aligning high-resolution remote sensing images with quantitative built environment metrics. We then evaluate both general-purpose and remote sensing VLMs (RS-VLMs) by applying VQAs to tens of built environment metrics across three complexity levels. Our results reveal that current state-of-the-art models consistently fall short on numeric reasoning tasks. We further conduct in-depth analyses across models, question types, and geographic locations, uncovering insights into performance variability and task-specific challenges.  

\end{abstract}    

\section{Introduction}

Quantitative reasoning is central to scientific discovery and real-world decision-making. While recent large language models (LLMs) and vision–language models (VLMs) have achieved strong performance in logical inference and multimodal understanding, their ability to perform quantitative reasoning grounded in built environments remains underexplored \cite{chow2025physbench}. This limitation is particularly critical as planning and policy decisions rely on accurate interpretation of spatial structure and numerical reasoning.

Quantitative reasoning can facilitate the analysis of built environment metrics, such as building density, land-use composition, and green coverage, revealing quantifiable geospatial patterns. Traditionally, these metrics are computed through time-consuming pipelines combining GIS, remote sensing imagery, and manual annotations, which require substantial domain expertise and limit scalability. In fact, high-resolution remote sensing imagery offers a scalable and structured visual representation of cities, naturally containing rich built environment and geospatial information. Quantitative reasoning via remote sensing imagery is an ideal testbed for grounded quantitative reasoning because it requires diverse reasoning capabilities including but not limited to object detection, proportional estimation, spatial reasoning, and compositional numerical inference.

Recent progress in remote sensing VLMs (RS-VLMs) \cite{kuckreja_geochat_2024, pang_vhm_2025, wang_geollava-8k_2025, yao2025falcon} and data benchmarks \cite{wang_xlrs-bench_2025, li_vrsbench_2024, zhou_urbench_2025} have demonstrated promising reasoning capabilities by combining remote sensing imagery and large VLMs. However, existing efforts primarily emphasize recognition and qualitative visual question answering (VQA), with limited focus on quantitative reasoning. While the RS-VLMs demonstrate strong performance in captioning and semantic understanding, their numeric reasoning capabilities remain underexplored, such as evaluating spatial contexts or calculating urban density metrics. 

In this work, we introduce USAI-Quant, a large-scale benchmark designed to systematically evaluate quantitative reasoning and extract built environment metrics from remote sensing imagery. The benchmark spans hundreds of U.S. cities and covers diverse urban forms, geographic regions, and reasoning complexities. Unlike other benchmarks, USAI-Quant focuses on numerically verifiable built environment metrics, enabling precise evaluation of reasoning accuracy beyond qualitative outputs.
The key contributions of this work are:

\begin{itemize}

\item \textbf{Data construction pipeline.} We design a scalable data preparation pipeline including data collection, urban feature extraction, VQA annotation, and human verification to ensure data quality and reliability.

\item \textbf{USAI-Quant benchmark.} As shown in Figure~\ref{fig:vqa_illustration}, we release a large-scale quantitative Urban and Spatial AI benchmark dataset, which covers hundreds of U.S. cities and provides diverse geographic contexts with varying reasoning complexity levels in VQA tasks.


\item \textbf{Comprehensive model benchmarking.} We benchmark both general-purpose and RS-specific large VLMs using built environment metrics, revealing their quantitative performance and key limitations. 

\item \textbf{Variability and robustness analysis.} We conduct an in-depth analysis across geographic locations, model scales, prompt design, reasoning complexities, and question formats, offering insights into how and why quantitative reasoning could fail in extracting built environment metrics. 


\end{itemize}

\section{Related Work}
\label{sec:related}

\subsection{Large Vision-Language Models}

The architecture of vision–language models (VLMs) typically consists of a visual encoder and a text decoder \cite{bordes2024introduction}. Advances in deep learning for computer vision and natural language processing, along with large-scale representation learning methods such as CLIP~\cite{radford_learning_2021}, have reshaped this paradigm by aligning image and text embeddings through extensive image–text pair training. The emergence of large language models (LLMs), including ChatGPT and LLaMA \cite{touvron2023llama}, further enhances text decoding and generation capabilities, enabling more sophisticated multimodal reasoning. Recent large VLMs have extended this framework toward comprehensive task solving rather than domain-specific applications. For example, Qwen \cite{bai2023qwen} extends a large language model with a vision encoder to enable unified multimodal understanding and structured visual reasoning, while InternVL \cite{chen_internvl_2024} focuses on scaling the visual backbone with native vision–language co-training to improve cross-modal alignment and long-context reasoning. Although these general-purpose VLMs perform well on open-domain visual question answering, they exhibit substantial limitations when tackling domain-specific quantitative reasoning tasks, particularly in urban analytics.

\subsection{Remote-Sensing Vision-Language Models}

The rapid evolution of remote-sensing vision–language models (RS-VLMs) is driven by the emergence of large-scale multimodal benchmarks, architectural advancements, and growing application demands. In particular, the rise of large language models has stimulated the development of remote-sensing foundation models \cite{hu_rsgpt_2025, feng_citygpt_2025, bazi_rs-llava_2024, wang_geollava-8k_2025, muhtar_lhrs-bot_2024}, accelerating progress in remote-sensing image understanding. For example, GeoChat \cite{kuckreja_geochat_2024} enables high-resolution, multi-turn conversational reasoning over both image-level and region-specific queries. VHM \cite{pang_vhm_2025} enhances multimodal alignment through detailed captioning and instruction tuning with factual and deceptive questions, promoting more reliable scene understanding. Falcon \cite{yao2025falcon} further improves performance across diverse benchmarks via large-scale instruction tuning tailored to remote sensing tasks. Despite these advances in semantic interpretation, quantitative reasoning capabilities remain insufficiently examined. Current models often answer quantitative questions at a recognition level, lacking robust arithmetic computation, proportional estimation, and multi-step numerical reasoning required for structured analytical tasks.

\subsection{Remote Sensing VQA Benchmarks}

To advance remote sensing models, a variety of benchmarks have been introduced to expand and diversify recognition tasks, thereby facilitating model development and enabling more comprehensive evaluation. These benchmarks progressively raise the standards for remote sensing imagery understanding, particularly in support of real-world applications. For example, RSVQA \cite{lobry2020rsvqa} established an early foundation for applying deep learning–based computer vision models to remote sensing VQA, incorporating tasks related to object recognition and image description within a question–answering framework. Building upon this foundation, more advanced and challenging datasets have been developed to support algorithmic evolution and robustness analysis. Representative VQA-oriented datasets include FloodNet\cite{rahnemoonfar2021floodnet}, RSVQA~\cite{lobry2020rsvqa}, RSIVQA~\cite{zheng2021mutual}, and CRSVQA~\cite{zhang2023multistep}, which vary in spatial resolution, scene complexity, and reasoning requirements. With the emergence of LLMs and large VLMs, recent benchmarks have further extended evaluation toward multimodal reasoning capabilities in remote sensing contexts \cite{wang_xlrs-bench_2025, zhou_urbench_2025, li_vrsbench_2024, danish2025geobench, kuckreja_geochat_2024, zi_rsvlm-qa_2025}. These studies highlight the limitations of general-purpose VLMs in handling domain-specific visual patterns, spatial reasoning, and quantitative inference in remote sensing imagery, while simultaneously driving progress in adapting multimodal models to geospatial applications. However, most existing benchmarks primarily emphasize qualitative assessment, while quantitative reasoning remains relatively superficial. In many cases, evaluation is limited to basic object counting or single-object size estimation, without addressing more complex urban metrics, multi-factor aggregation, and higher-level numerical inference.

\subsection{Built Environment Metrics}
Traditionally, researchers have developed a large number of quantitative built environment metrics to understand urban systems. Early work emphasized a 3D framework, including density, diversity, and design, as key dimensions characterizing the built environment \citep{cervero_travel_1997}. Subsequent work expanded such metrics to include accessibility, transportation proximity, and other spatial network properties \citep{ewing_travel_2010, boeing_osmnx_2017}. The built environment metrics are increasingly used to explain patterns in walking, public transit, and car usage to improve transportation sustainability \citep{aditjandra_exploring_2016, farahani_review_2013}. Such metrics can also reveal spatial inequalities in reaching opportunities, especially for vulnerable groups \citep{moreno_introducing_2021, weng_15-minute_2019, capasso_da_silva_accessibility_2020}. Some researchers link built environment patterns to urban vitality \citep{lyu_investigating_2025}, crime risk \citep{kim_density_2021}, and social equity concerns \citep{gu_density_2022}. These findings indicate that the quantitative built environment metrics serve as both diagnostic and intervention tools for improving cities. 
Recently, VLMs have facilitated the understanding of the built environment by enabling higher-level semantic understanding, cross-modal reasoning, and qualitative VQA tasks, thus reducing reliance on labor-intensive human interpretation or narrowly designed computational pipelines. However, it remains an open research question how effective VLMs are in delivering accurate and reliable quantitative built environment metrics.

\section{USAI-Quant}

\begin{figure*}[h!]
    \centering
    \includegraphics[width=1\linewidth]{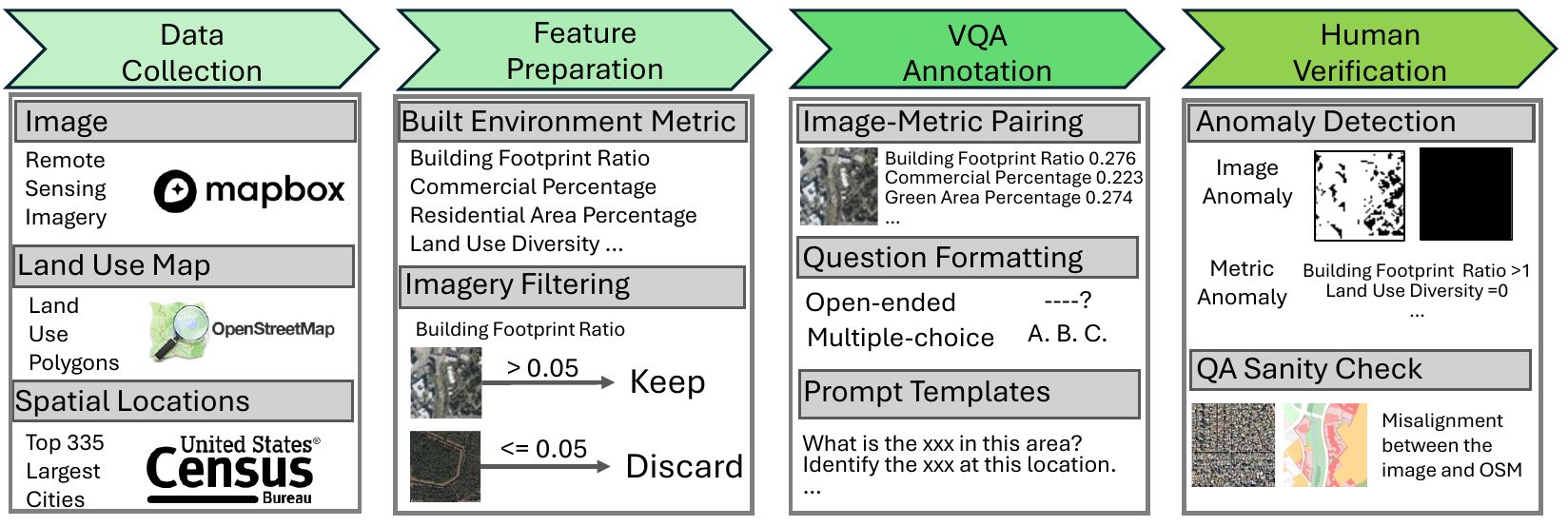}
    \caption{Benchmark preparation pipeline.}
    \label{fig:pipeline}
\end{figure*}

We prepared USAI-Quant following standard data preparation procedures, including data collection, feature engineering, visual question–answer generation, and human verification, as illustrated in Figure~\ref{fig:pipeline}.

\subsection{Data Collection}
We collected two primary data sources from established platforms: remote sensing imagery from Mapbox \cite{rzeszewski2023mapbox} and land use spatial statistics from OpenStreetMap (OSM) \cite{OpenStreetMap}. Both datasets were aggregated and organized according to city size, based on Gazetteer place-level national statistics provided by the U.S. Census Bureau.

\textbf{Metropolitan area selections.} City center coordinates were obtained from the U.S. Census Bureau, specifically from the U.S. Gazetteer Files. We selected 335 largest cities in descending order of metropolitan statistical area size as defined by the U.S. Census Bureau. These cities exhibit rich and diverse urban footprints, enabling the reliable derivation of informative built environment metrics for urban planning and decision-making. Moreover, their dense and spatially heterogeneous built environments present greater complexity, making them both more challenging and more critical for assessing the competence of vision–language models.

\begin{figure}[h!]
    \centering
    \includegraphics[width=0.95\linewidth]{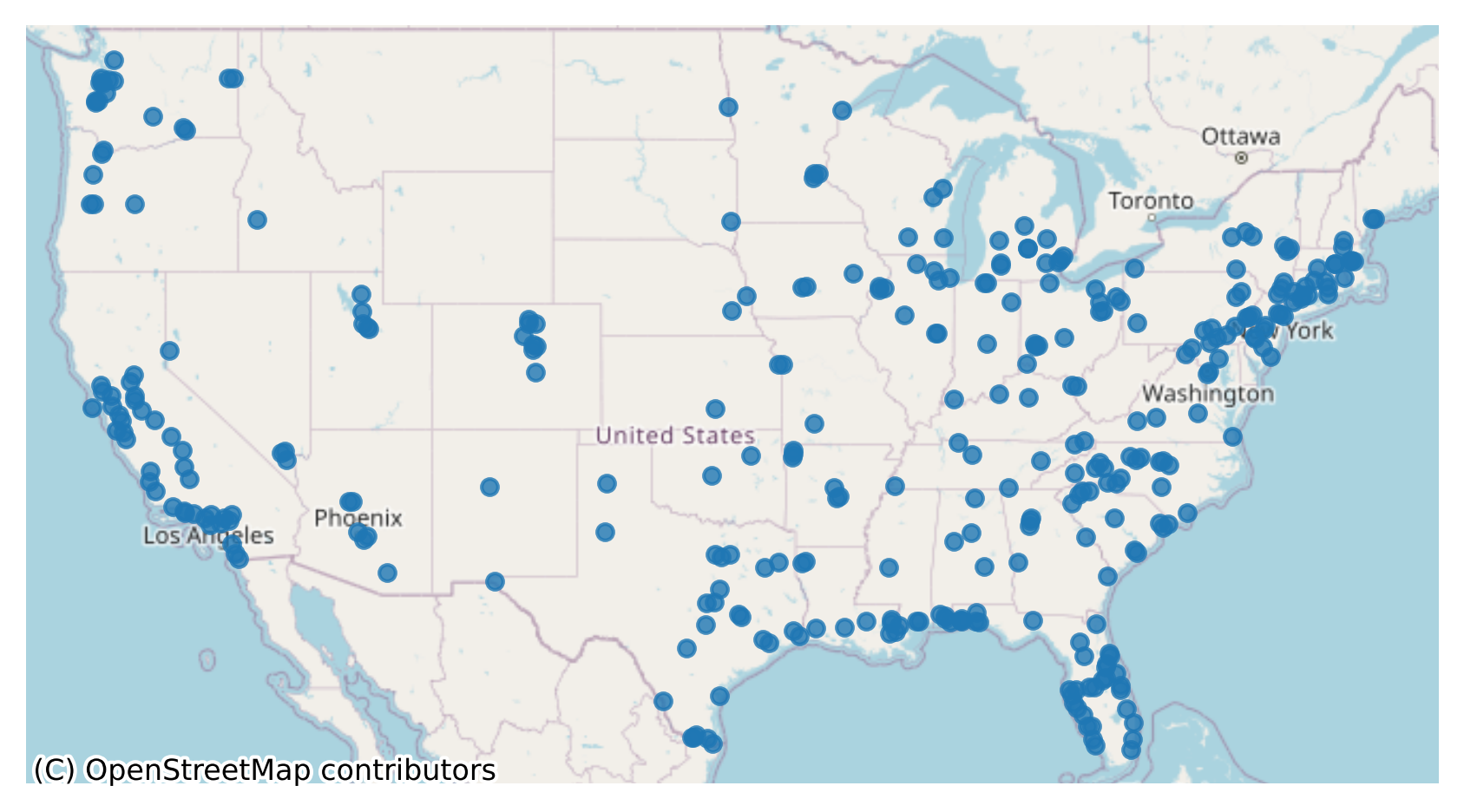}
    \caption{Geographic locations of the 335 selected metropolitan areas used for urban imagery analysis.}
    \label{fig:cities_illustration}
\end{figure}

\textbf{Remote sensing imagery.} According to the selected 335 largest metropolitan areas, we perform remote sensing imagery extraction. For each city, a bounding box is defined to constrain the study region, and imagery is extracted from Mapbox  at zoom level 17, corresponding to a spatial resolution of 672 × 672 pixels. Each image covers an approximate ground area of 450 m × 450 m. Given the city center coordinates, we generate a 4 × 4 grid of neighboring tiles and collect up to 16 images per city, capturing the urban environment surrounding the central area.

\textbf{Land use spatial statistics.} Land use planning maps are obtained from OSM, which provides detailed land use polygon annotations. For our analysis, under the land use tag, we select the categories residential, commercial, industrial, and green-related classes, including greenery, greenfield, grass, forest, and farmland. Under the natural tag, we select water. Based on these polygons and their spatial statistics, a set of built environment metrics is derived to characterize the structural and functional composition of each city.



\begin{table*}[ht]
    \small
    \centering
    \caption{
        Comparing state‑of‑the‑art remote sensing VQA benchmarks with respect to both general performance criteria and urban‑focused evaluation metrics. \protect\manual{} indicates that no questions correspond to this feature, 
\protect\partialmanual{} indicates that some questions are partially relevant to this feature, and 
\protect\machine{} indicates that all questions focus on this feature.
    }
    \label{tab:rs_bench}
    \renewcommand{\arraystretch}{1.2}
    \setlength{\tabcolsep}{3.5pt}
    \begin{tabular}{l|c|cccccc}
        \hline
        \textbf{Benchmark} & \textbf{Year} & \textbf{Images} & \textbf{Average Resolution} & \textbf{Spatial Unit Area} &\textbf{Questions} & \textbf{Urban Questions} & \textbf{Arithmetic Reasoning} \\
        \hline
         GeoChat-Bench~\cite{kuckreja_geochat_2024}        & 2024 & 105K & 800$\times$800 & 92$\times$92m$^2$ & 306K &  \partialmanual & \manual   \\
         VRSBench~\cite{li_vrsbench_2024}   & 2024 & 29K & 512$\times$512 & 410$\times$410m$^2$ & 123K &  \partialmanual & \manual \\
         UrBench~\cite{zhou_urbench_2025}   & 2025 & 4.2K & 640$\times$640 &  200$\times$200m$^2$ &  11.6K & \machine & \manual \\
         RSVLM-QA~\cite{zi_rsvlm-qa_2025}   & 2025 & 14K & 512$\times$512 & 1,400$\times$1,400m$^2$ &  162K & \partialmanual  & \manual \\
         XLRS-Bench~\cite{wang_xlrs-bench_2025} & 2025 & 1.4K & 8,500$\times$8,500 & 5,000$\times$5,000m$^2$ & 46K & \partialmanual  & \manual  \\
         \hline
         USAI-Quant (ours)~&  2026 & 100K & 672$\times$672 & 450$\times$450m$^2$ & 699K &  \machine & \machine   \\
        %
        \hline
    \end{tabular}
\end{table*}

\subsection{Feature Preparation}

We prepare features for image–question pairs and corresponding metric answers by aligning OSM polygons with Mapbox remote sensing imagery. Using OSM, we extract spatial land use information, which is then leveraged to compute various built environment metrics. Based on metric calculations and preliminary image analysis, we filter and retain high-quality, informative imagery, producing reliable image–metric pairs for VQA reasoning tasks.

\textbf{Built environment metrics.} We selected 13 critical urban planning infrastructure metrics and formulated them as quantitative reasoning questions. These built environment metrics are listed in Table~\ref{tab:urban_metrics_summary}. Careful filtering of remote sensing imagery is essential to ensure that the derived built environment metrics are valid and accurate. To maintain data quality, building footprints were used as the primary criterion to screen and retain images containing sufficient building presence for reliable human footprint recognition and analysis. Based on the extracted imagery and land use data, we computed 13 built environment metrics that play prominent roles in urban planning and decision-making processes.

\textbf{Imagery filtering.} We perform imagery quality control to remove invalid or non-informative tiles, ensuring the reliability of downstream urban metric estimation by VQA. Inferior or invalid imagery refers to image regions that lack sufficient visual information to support reliable urban metric estimation. These could include tiles with no visible building footprints, homogeneous land-cover regions (e.g., entirely vegetation, water, or pavement), and blank or near-blank images caused by missing data. Such samples do not contain meaningful built-environment features and therefore cannot support quantitative reasoning about urban structure. Specifically, we use the building footprint ratio as a threshold: images with a footprint ratio of zero are excluded, as they provide insufficient information to support meaningful urban reasoning tasks

\begin{table*}[t]
\small
\centering
\caption{Selected built environment metrics \cite{santhanam2022quantification, oliveira2020urban, ewing2010travel} with definitions, explanation and reasoning levels.}
\label{tab:urban_metrics_summary}
\setlength{\tabcolsep}{3pt}
\renewcommand{\arraystretch}{1.05}
\begin{tabular}{p{3.8cm} >{\centering\arraybackslash}p{3.5cm} p{8cm} >{\centering\arraybackslash}p{0.6cm}}
\toprule
\textbf{Metric} & \textbf{Formula} & \textbf{Description} & \textbf{Level} \\
\midrule

Building Footprint Ratio &
$\sum A_i^{\text{footprint}} / A_{\text{total}}$ &
Share of land covered by buildings; development intensity. & L1 \\

Commercial Area Percentage &
$A_{\text{commercial}} / A_{\text{total}}$ &
Proportion of commercial land. & L1 \\

Green Area Percentage &
$A_{\text{green}} / A_{\text{total}}$ &
Vegetated land share; ecological capacity. & L1 \\

Residential Area Percentage &
$A_{\text{residential}} / A_{\text{total}}$ &
Housing land proportion. & L1 \\

Urban Blue Area Percentage &
$A_{\text{blue}} / A_{\text{total}}$ &
Surface water proportion. & L1 \\

Industrial Area Percentage &
$A_{\text{industrial}} / A_{\text{total}}$ &
Industrial land share. & L1 \\

Impervious Space Ratio &
$A_{\text{impervious}} / A_{\text{total}}$ &
Fraction of impermeable surfaces. & L1 \\

Amenity Density &
$N_{\text{amenities}} / A_{\text{total}}$ &
Amenities per unit area; accessibility. & L2 \\

Unique Land Uses &
$K$ &
Number of distinct land use types. & L2 \\

Blue/Built Ratio &
$A_{\text{blue}} / A_{\text{built}}$ &
Water relative to built-up land. & L2 \\

Green–Blue Index &
$A_{\text{green}} / A_{\text{blue}}$ &
Vegetation relative to water. & L2 \\

Blue–Green Ratio &
$A_{\text{blue}} / A_{\text{green}}$ &
Water relative to vegetation. & L2 \\

Land Use Diversity &
$-(1/\ln K)\sum p_k \ln p_k$ &
Normalized entropy of land use distribution. & L3 \\

\bottomrule
\end{tabular}
\end{table*}

\subsection{VQA Annotation}

We categorize questions into three reasoning levels based on the number and complexity of inference steps required to derive the answer. 

\textbf{Question types.} We prepare two categories of question sets for VQA benchmarking. The first consists of open-ended questions, in which each question directly requests a numerical answer. The second comprises closed-form multiple-choice questions, where the model selects the correct answer from several candidate options. All candidate options are numerical values. These options are generated based on quantile intervals: in addition to the ground-truth value, three alternative numerical candidates are randoly sampled from other quantile ranges to serve as distractors.

\textbf{Reasoning level.} We develop a reasoning system. Level~1 metrics rely on direct visual recognition of land-cover and built-form elements, enabling region identification and area estimation without complex relational inference. As shown in Table~\ref{tab:urban_metrics_summary}, these metrics are obtained by aggregating visually distinguishable regions such as buildings, vegetation, and paved surfaces. Level~2 metrics require relational reasoning and arithmetic operations across multiple regions or attributes, such as computing spatial ratios (Table~\ref{tab:urban_metrics_summary}). This level evaluates intermediate numerical reasoning beyond single-region recognition. Level~3 metrics demand multi-step compositional reasoning that integrates multiple land use categories and their proportions into a unified inference. Represented by Land Use Diversity in our benchmark, this level tests holistic scene understanding and structured reasoning over complex urban systems.

\textbf{Prompt engineering.} We design 10 diverse prompt templates to query the same underlying urban metric as shown in Table~\ref{tab:prompt_templates}. This prompt diversity evaluates whether the tested VLM demonstrates robust question understanding beyond surface-level phrasing, thereby enabling a more reliable and rigorous assessment of its ability to infer the intended metrics. In addition, these templates allow us to analyze the uncertainty and variability of the model’s generated responses under semantically equivalent queries.

\begin{table}[h!]
\small
\centering
\caption{Alternative prompt templates using [XXX] as a placeholder for built environment metrics.}
\begin{tabular}{cl}
\hline
\textbf{\#} & \textbf{Prompt Template} \\
\hline
1 & What is the [XXX] in this area? \\
2 & Can you describe the [XXX] at this location? \\
3 & Please indicate the [XXX] for this location. \\
4 & Could you provide the [XXX] value observed here? \\
5 & Identify the [XXX] at this location. \\
6 & Determine the [XXX] across this region. \\
7 & Explain the [XXX] measured here. \\
8 & Tell me the [XXX] characteristic of this area. \\
9 & What can you tell about the [XXX] here? \\
10 & Provide the [XXX] found in this region. \\
\hline
\end{tabular}
\label{tab:prompt_templates}
\end{table}

\subsection{Human Verification}

We performed verification through anomaly detection and sanity checks to ensure data integrity and validity. Specifically, all percentages and ratios were constrained within the valid range of 0 to 1, and land use counts were verified not to exceed the predefined category limits. Images containing undefined or irrelevant land use categories were excluded to maintain benchmark reliability. All remote sensing imagery was inspected to ensure clear building footprints without blank regions or distorted features. Additionally, OSM polygons were validated by overlaying them with corresponding remote sensing imagery to confirm spatial and planning consistency, thereby ensuring reliable question–answer generation and evaluation.

\begin{table}[h]
\centering
\footnotesize
\caption{Overall statistics of the USAI-Quant.}
\label{tab:USAI-Quant_statistics}
\begin{tabular}{lr @{\hspace{10pt}} lr}
\toprule
\textbf{Metric} & \textbf{Value} & \textbf{Metric} & \textbf{Value} \\
\midrule
Total Images & 100,523 
& Training Images & 95,523 \\

Total VQA Pairs & 699,341 
& Training VQA Pairs & 569,341 \\

Metrics per Image & 13 
& Testing Images & 5,000 \\

Question Length (words) & 9.23 
& Testing VQA Pairs & 130,000\\
\bottomrule
\end{tabular}
\end{table}

\subsection{Dataset Characteristics Summary}

\begin{figure}[h]
    \centering
    \includegraphics[width=0.85\linewidth]{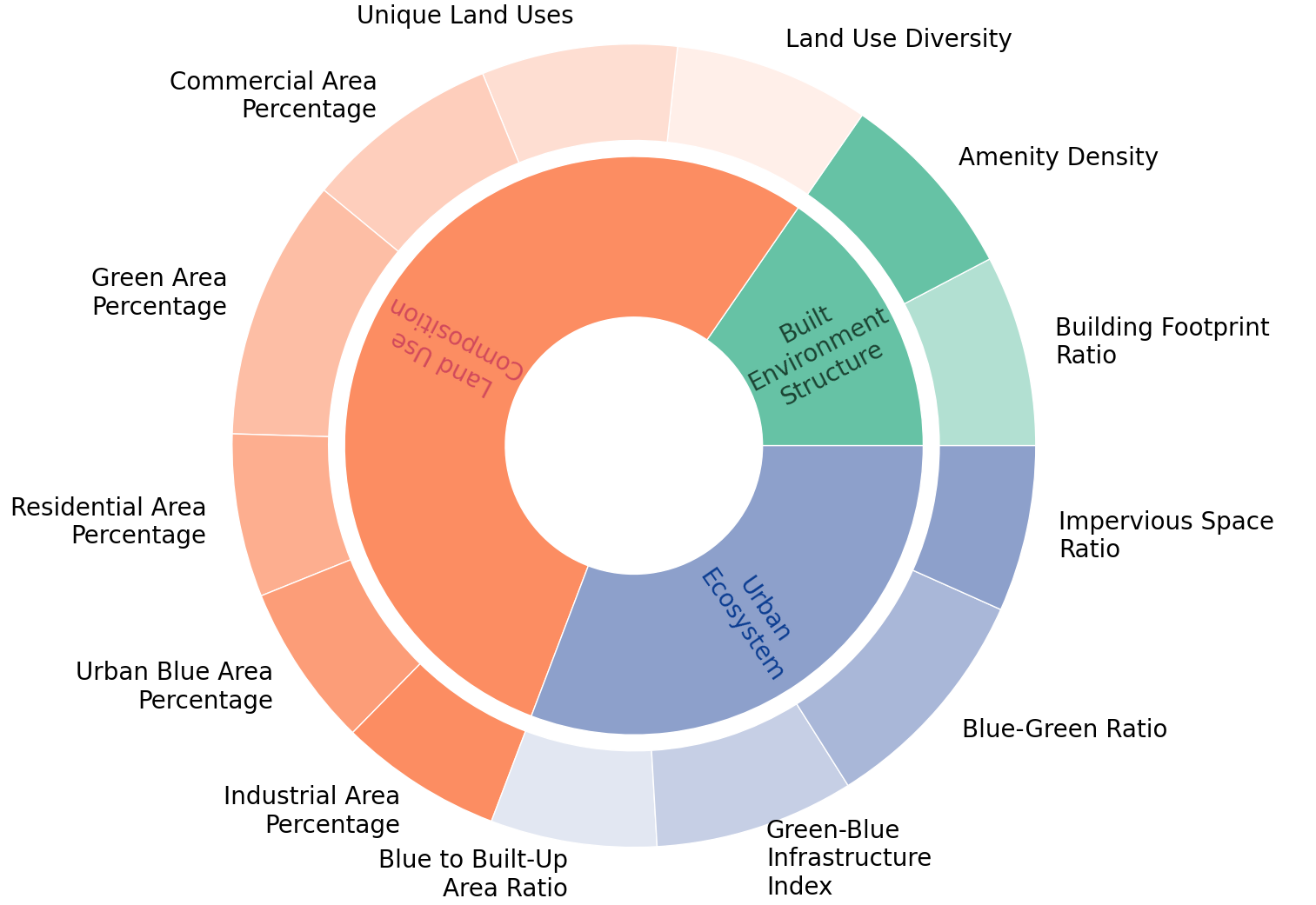}
    \caption{Different built environment metric categories.}
    \label{fig:question_structures}
\end{figure}

We summarize the dataset structure in Figure~\ref{fig:question_structures}. The 13 metrics are grouped into three major categories: land use composition, urban ecosystem, and built environment structure, all of which are critical for urban planning. Land use composition constitutes the largest component, reflecting its central role in assessing land use patterns for planning decisions. All questions are provided in both open-ended and multiple-choice formats, as illustrated in Figure~\ref{fig:vqa_illustration}.

We also display the summary of USAI-Quant in Table~\ref{tab:USAI-Quant_statistics}. Note that in the training set, each image has up to 26 questions covering all metrics, and all test set images include all 26 questions. To balance the training data, we retained all non-zero metric questions while reducing the number of zero-metric questions for each image.

\section{Experiments}
\subsection{Models and Metrics}


We selected a diverse set of state-of-the-art VLMs to benchmark their performance on urban imagery reasoning tasks. The general-purpose models include LLaVA-1.6-7B \cite{liu_improved_2024}, InternVL-3-8B \cite{chen2024internvl}, QwenVL-3-8B \cite{bai2025qwen3}, and MGM-7B \cite{li2025mini} while domain-specific models include GeoChat-7B \cite{kuckreja_geochat_2024}, VHM-7B \cite{pang_vhm_2025}, and Falcon-7B \cite{yao2025falcon}, covering a range of architectures, training paradigms, and multi-modal capabilities. This selection spans models optimized for instruction-following, spatial reasoning, geospatial imagery interpretation, and general multi-step visual reasoning. By evaluating these models, we aim to assess both direct visual recognition tasks corresponding to simpler metrics (L1) and more complex relational or compositional reasoning tasks (L2 and L3), enabling a comprehensive analysis of robustness, uncertainty, and generalization in measuring urban planning and ecosystem metrics.

For open-ended questions, we evaluate performance using both success rate and accuracy with a tolerance range. The success rate is defined based on answer format: if the model returns a numeric value, it is counted as a successful response. Since all questions require a valid number, this metric primarily reflects prompt effectiveness. In addition, we assess accuracy by comparing model outputs with ground-truth values. Correctness is determined within predefined tolerance thresholds: $\pm 0.1$ for ratios or percentages, $\pm 1$ for integer values, and $\pm 0.2$ for land-use diversity.

\subsection{Performance Results}

We compare the selected models across different built environment metrics in terms of accuracy. The key observations are summarized as follows.

\begin{table*}[t]
\centering
\small
\setlength{\tabcolsep}{2.5pt}
\renewcommand{\arraystretch}{1.0}
\caption{Quantitative reasoning performance across built environment metrics. Each cell reports success rate (\%) / numeric tolerance accuracy (\%). Metrics are sorted by reasoning level.}
\label{tab:quant_reasoning_sorted}
\begin{tabular}{l c|cccc|ccc}
\hline
\multirow{2}{*}{Built Environment Metrics} & \multirow{2}{*}{Level} & \multicolumn{4}{c|}{\textbf{General-Purpose Models}} & \multicolumn{3}{c}{\textbf{Domain-Specific Models}} \\
\cline{3-9}
 & & LLaVA-1.6 & InternVL-3 & QwenVL-3 & MGM & GeoChat & VHM & Falcon \\
\hline

Building Footprint Ratio & L1 & 78.5 / 5.1 & 49.4 / 4.1 & 41.9 / 2.1 & 47.9 / 5.9 & 97.8 / 12.7 & 93.5 / 8.6 & 93.8 / 10.2 \\
Commercial Area Percentage & L1 & 79.3 / 8.4 & 57.5 / 4.1 & 62.6 / 3.0 & 77.1 / 4.2 & 95.9 / 10.2 & 92.9 / 12.2 & 92.7 / 10.3 \\
Green Area Percentage & L1 & 76.8 / 20.3 & 61.6 / 16.3 & 59.6 / 14.7 & 83.9 / 17.1 & 90.9 / 35.8 & 97.8 / 30.7 & 92.1 / 29.5 \\
Residential Area Percentage & L1 & 78.6 / 9.3 & 58.6 / 5.2 & 57.2 / 2.9 & 76.4 / 6.1 & 94.5 / 12.4 & 90.7 / 9.7 & 91.5 / 9.6 \\
Urban Blue Area Percentage & L1 & 82.6 / 15.4 & 60.9 / 14.3 & 57.7 / 8.0 & 81.2 / 13.7 & 94.8 / 18.8 & 95.5 / 17.1 & 90.9 / 20.4 \\
Industrial Area Percentage & L1 & 81.1 / 5.1 & 61.4 / 3.9 & 62.3 / 2.0 & 80.8 / 2.9 & 95.4 / 14.3 & 97.1 / 10.2 & 92.0 / 14.4 \\
Impervious Space Ratio & L1 & 73.7 / 7.9 & 69.0 / 9.4 & 66.0 / 6.9 & 73.0 / 4.8 & 91.0 / 14.5 & 89.0 / 12.4 & 92.0 / 12.2 \\

Amenity Density & L2 & 51.1 / 6.1 & 38.1 / 5.1 & 39.6 / 1.0 & 50.4 / 5.7 & 34.3 / 9.3 & 31.0 / 10.3 & 35.9 / 9.5 \\
Unique Land Uses & L2 & 19.6 / 8.1 & 20.1 / 4.1 & 14.5 / 3.1 & 15.7 / 5.1 & 14.8 / 8.7 & 14.0 / 10.5 & 13.0 / 9.4 \\
Blue to Built-Up Area Ratio & L2 & 76.3 / 6.8 & 69.0 / 5.8 & 66.0 / 2.9 & 59.0 / 2.9 & 97.0 / 13.2 & 94.0 / 10.7 & 96.0 / 12.3 \\
Green-Blue Infrastructure Index & L2 & 51.2 / 6.1 & 49.0 / 6.2 & 44.0 / 3.9 & 72.0 / 4.2 & 50.0 / 11.7 & 61.0 / 12.4 & 59.0 / 14.6 \\
Blue-Green Ratio & L2 & 74.5 / 8.8 & 71.0 / 7.9 & 64.0 / 7.1 & 68.0 / 5.8 & 92.0 / 12.2 & 88.0 / 9.8 & 94.0 / 12.1 \\

Land Use Diversity & L3 & 9.8 / 9.7 & 9.6 / 6.1 & 10.7 / 2.1 & 10.2 / 4.1 & 14.4 / 13.6 & 14.6 / 9.7 & 10.3 / 9.1 \\

\hline
\end{tabular}
\end{table*}

\begin{table*}
\centering
\small
\setlength{\tabcolsep}{5pt} 
    \caption{
        Comparison of quantitative reasoning accuracy (\%) across different state-of-the-art large VLMs in multiple-choice question. 
    }
\begin{tabular}{l c|cccc|ccc}
\hline
\multirow{2}{*}{Built Environment Metrics} & \multirow{2}{*}{Level} & \multicolumn{4}{c|}{\textbf{General-Purpose Models}} & \multicolumn{3}{c}{\textbf{Domain-Specific Models}} \\
\cline{3-9}
&  & LLaVA-1.6 & InternVL-3 & QwenVL-3 & MGM & GeoChat & VHM & Falcon  \\
\hline
Building Footprint Ratio & L1 & 27.71 & 36.21 & 35.21 & 33.41 & 22.14 & 23.09 & 25.55  \\
Commercial Area Percentage & L1 & 27.19 & 37.05 & 35.94 & 32.87 & 22.48 & 23.62 & 26.21  \\
Green Area Percentage & L1 & 35.06 & 45.72 & 44.18 & 43.96 & 41.29 & 42.97 & 42.01  \\
Residential Area Percentage & L1 & 25.84 & 36.90 & 35.63 & 34.25 & 22.71 & 23.02 & 24.89  \\
Urban Blue Area Percentage & L1 & 24.97 & 34.88 & 33.94 & 32.36 & 21.85 & 23.66 & 26.02  \\
Industrial Area Percentage & L1 & 26.68 & 36.47 & 34.59 & 33.72 & 22.94 & 21.28 & 25.33  \\
Impervious Space Ratio & L1 & 25.94 & 36.02 & 35.77 & 34.19 & 22.88 & 23.14 & 25.09  \\
Amenity Density & L2 & 26.41 & 35.94 & 36.12 & 35.57 & 21.80 & 20.76 & 27.73  \\
Unique Land Uses & L2 & 26.52 & 35.13 & 36.09 & 34.08 & 21.64 & 23.41 & 25.88  \\
Blue to Built-Up Area Ratio & L2 & 25.02 & 35.61 & 36.41 & 32.79 & 21.56 & 22.84 & 26.57  \\
Green-Blue Infrastructure Index & L2 & 25.23 & 37.12 & 35.08 & 33.58 & 22.26 & 22.95 & 24.91  \\
Blue-Green Ratio & L2 & 24.66 & 35.28 & 34.47 & 32.63 & 23.41 & 22.89 & 26.85  \\
Land Use Diversity & L3 & 24.88 & 36.84 & 34.62 & 32.91 & 23.02 & 25.45 & 24.77 \\
\hline
\end{tabular}
\label{tab:sota_accuracy}
\end{table*}

\textbf{Observations on open-ended questions.} The success rate for numeric answers is defined as the proportion of cases where at least one number within a reasonable range is returned. We observe that Built Environment Metrics expressed as percentages or ratios achieve higher success rates. This is likely because the language decoder can more easily associate these terms with clear semantic meanings. With well-defined concepts, language models are able to return numeric answers more reliably. Land use diversity is also challenging for all VLMs, likely because diversity is not well captured in their semantic representations, suggesting that performance could be improved through better fine-tuning or prompt design.

\textbf{Observations on multiple-choice questions.} Compared to open-ended questions, multiple-choice questions show greater performance improvements across general-purpose VLMs. This suggests that comparative reasoning is already partially captured in these models, whereas domain-specific VLMs still struggle when comparative candidates are involved. Therefore, further reasoning-focused training remains important for domain-specific models.

\textbf{Observations on prompt words.} Green Area Percentage and Urban Blue Area Percentage consistently achieve the best performance. This is likely because ``green'' and ``blue'' are well-defined in the vision-language representation, and ``percentage'' provides a clear semantic target for the language decoder. Moreover, the image encoder can more easily perceive and estimate green or blue areas through detection or segmentation. Unique Land Uses and land-use diversity remain challenging for all VLMs in open-ended questions, likely because the prompts require explicit quantitative reasoning that is not well embedded in these models. However, the performance gap narrows when the questions are reformulated as multiple-choice prompts, where numerical information and answer candidates are more clearly specified.

\textbf{Comparison between general-purpose and domain-specific models.} For open-ended questions, domain-specific models can return numeric answers more successfully than general-purpose models, indicating that instruction tuning on remote sensing imagery recognition and prompt design is effective. However, the improvement is less significant when answers are evaluated numerically against ground truth. This pattern further changes with multiple-choice questions. Domain-specific models still exhibit weak logical reasoning and fail to leverage candidate choices effectively, whereas general-purpose models demonstrate stronger quantitative reasoning, allowing them to use multiple-choice options to boost performance.

\textbf{Observations on reasoning levels.} As expected, accuracy varies according to reasoning level: higher reasoning levels correspond to lower accuracy. This suggests that questions requiring more complex reasoning demand further optimization of language understanding and reasoning capabilities. Techniques such as chain-of-thought prompting could serve as a promising approach to improve performance on multi-step reasoning tasks.

\subsection{Performance Variability Analysis}

We further conduct a performance variability analysis to  evaluate the behavior of current VLMs on USAI-Quant. Specifically, we examine performance variations across model scales, prompt templates, and geographic locations on multiple-choice VQA tasks.

\begin{figure}[h]
    \centering
    \includegraphics[width=0.95\linewidth]{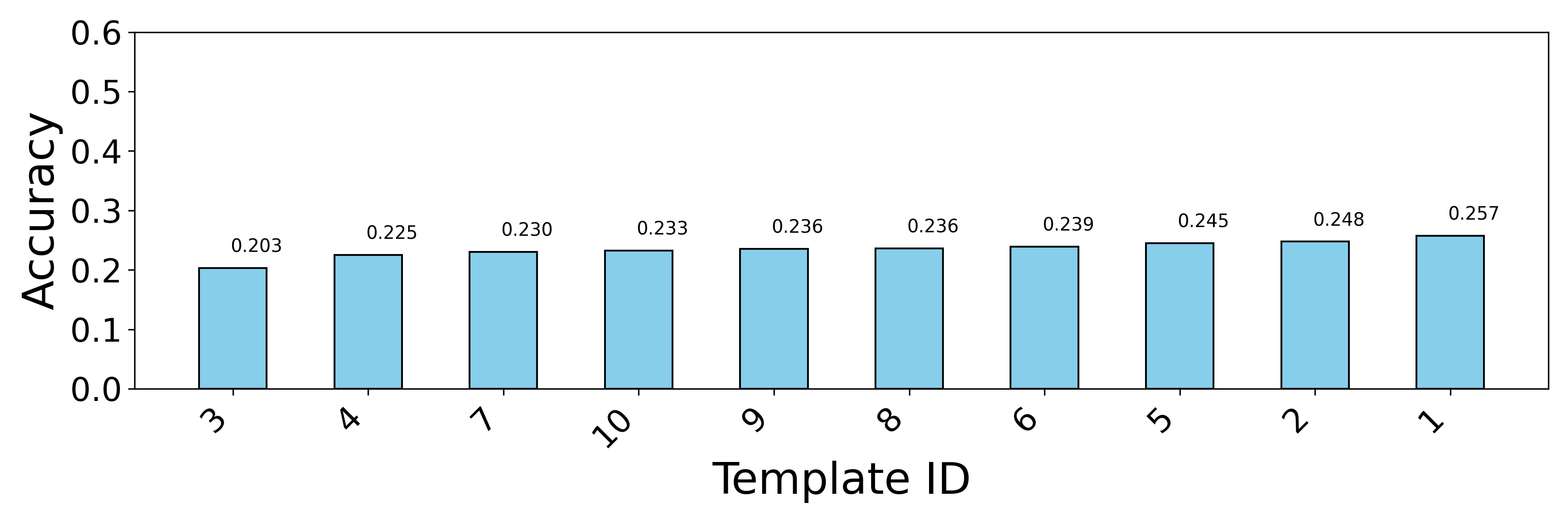}
    \caption{Accuracy of Qwen3-VL-8B across 10 templates.}
    \label{fig:accuracy_template}
\end{figure}

\textbf{Accuracy across question templates.} We evaluate performance variability across different question templates. As shown in Figure~\ref{fig:accuracy_template}, the effect of template variation on Qwen is minimal. These results suggest that paraphrasing contributes to robust training in general-purpose vision-language models, while simple prompt rewording has only a limited effect on performance.

\begin{figure}[h]
    \centering
    \includegraphics[width=0.85\linewidth]{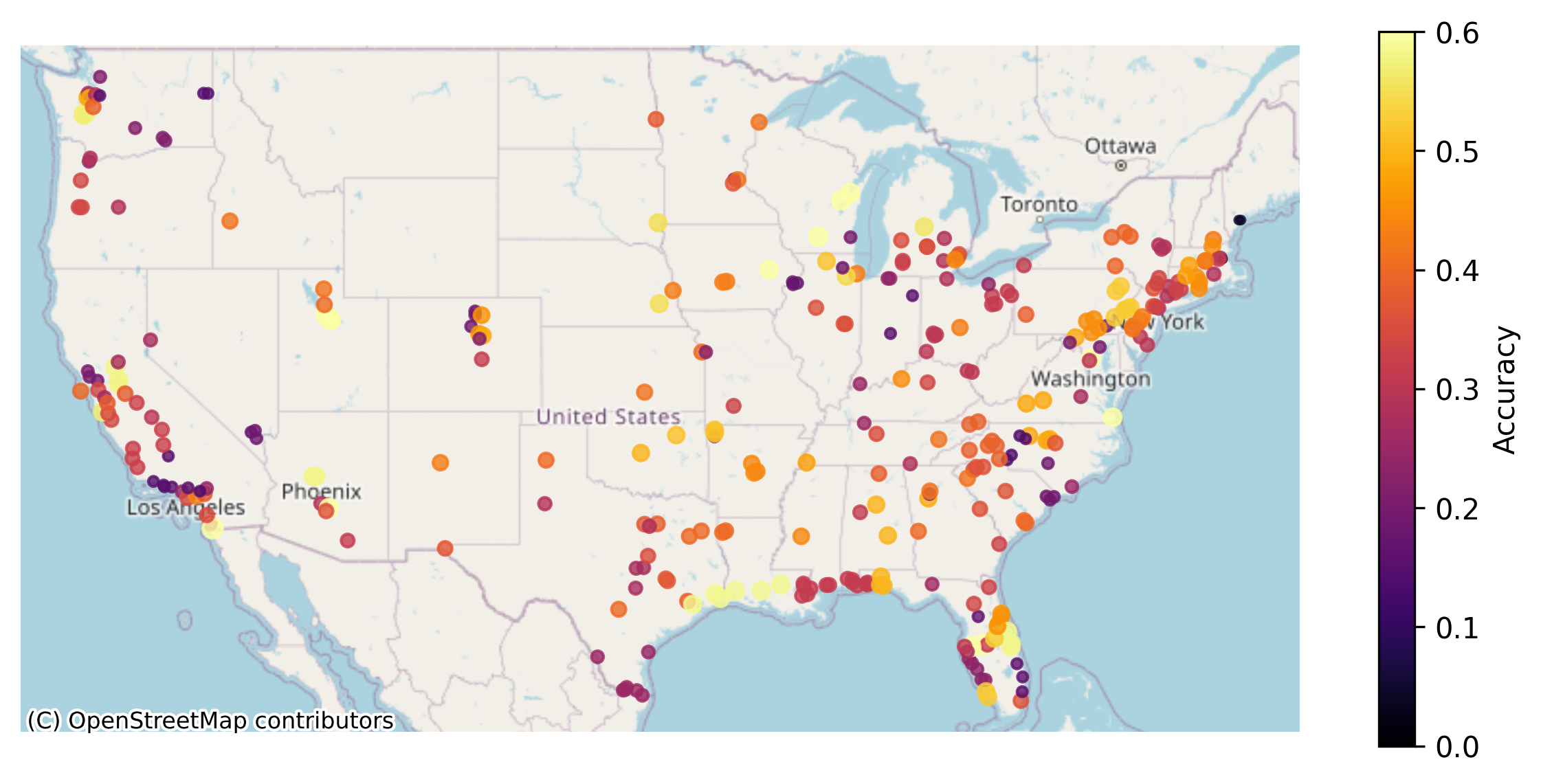}
    \caption{Accuracy of Qwen3-VL-8B across cities.}
    \label{fig:accuracy_spatial_distribution}
\end{figure}

\textbf{Accuracy across geographical locations.} Figure~\ref{fig:accuracy_spatial_distribution} shows accuracy across cities, highlighting variability between locations. The results indicate that, although some larger cities show higher accuracy, no consistent trend emerges across city sizes. We speculate this spatial variation may result from incomplete or lower-quality land use information in OSM planning data We observe the cities with lower accuracy usually shows lower quality of OSM data which introduces noise into derived metrics and weakens the connection between remote sensing data and urban indicators, potentially explaining the observed performance degradation in some cities. Improving OSM data is therefore crucial for enhancing vision–language models in Built Environment Metrics reasoning. Moreover, any planning or analysis relying on OSM data should account for these limitations and prioritize data quality improvements.

\textbf{Accuracy by model variants.} As model size increases up to 8B for Qwen3-VL or 7B for InternVL-3 parameters, performance improves accordingly. Beyond 8B for Qwen3-VL or 7B for InternVL-3, however, further scaling yields only marginal or negligible gains. This suggests that general reasoning ability largely drives question-answering performance below 8B for Qwen3-VL or 7B for InternVL-3. With additional architectural scaling or more pretraining data, the current benchmark shows minimal improvement in reasoning for domain-specific quantitative reasoning. Strengthening domain-specific capabilities will require focused improvements in the model’s ability to perform inference and reasoning grounded in domain knowledge.

\begin{figure}[h]
    \centering
    \includegraphics[width=0.85\linewidth]{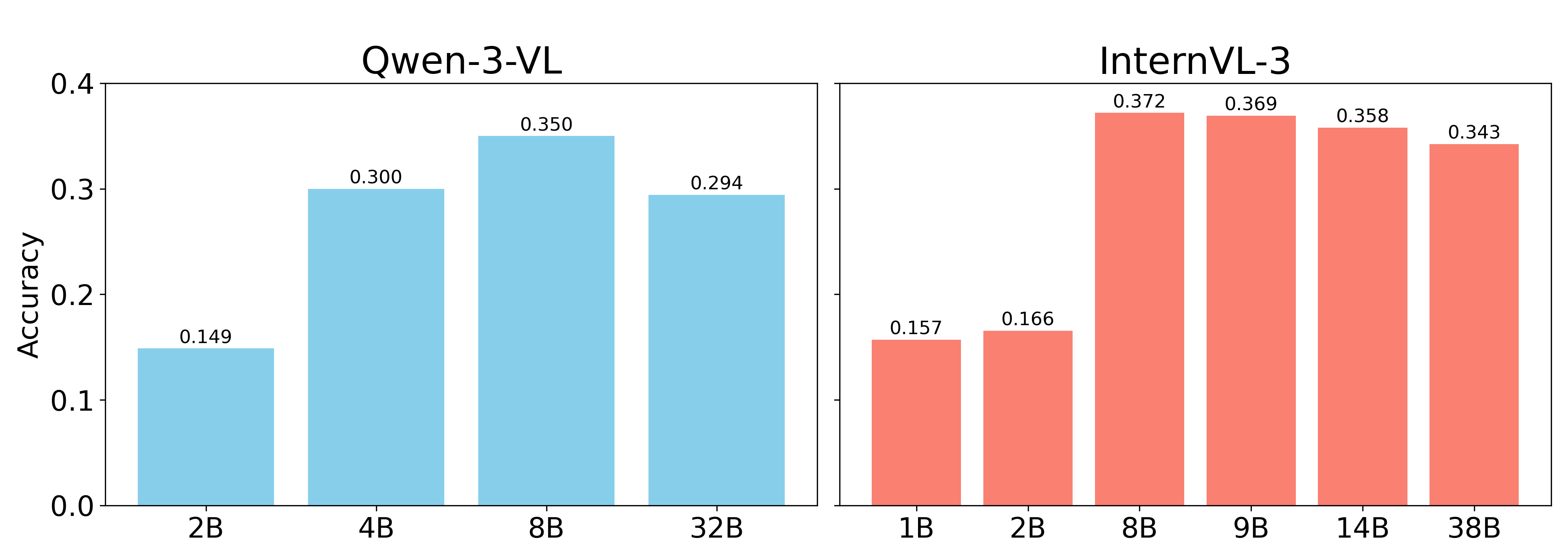}
    \caption{Effect of model size on Qwen3-VL and InternVL-3.}
    \label{fig:acc_model_size}
\end{figure}

\section{Conclusions}

This work develops a benchmark to evaluate the current state-of-the-art large VLMs on quantitative visual question answering for remote sensing imagery. The benchmark includes a larger volume of images and questions, emphasizing more challenging yet practically important quantitative tasks for urban planning applications. Based on the designed open-ended and closed-form question sets, our results indicate that general-purpose VLMs demonstrate limited remote sensing recognition capability. More importantly, several models exhibit insufficient quantitative reasoning ability and struggle to produce accurate numerical answers. In comparison, remote sensing domain-specific models show relatively stronger numerical response capability; however, their predictions still deviate considerably from the ground truth. Open-ended questions reveal greater limitations than closed-form multiple-choice questions. Although multiple-choice settings yield improved performance, overall accuracy remains below 50$\%$, indicating substantial room for improvement. We also observe that model responses tend to rely heavily on prompt keywords rather than robust visual–numerical understanding. Furthermore, we conduct a comprehensive analysis of performance variability across model variants, question templates, reasoning levels, and geographic locations. These findings provide more reliable references for future algorithm design and optimization of large VLMs for remote sensing visual question answering.

{
    \small
    \bibliographystyle{ieeenat_fullname}
    \bibliography{main}
}


\end{document}